# Parsing Akkadian Verbs with Prolog

Aaron Macks

Department of Computer Science
Brandeis University
Waltham, MA. 02454
`aaronm@cs.brandeis.edu`

## Abstract

This paper describes a parsing/generation system for finite verbal forms in Akkadian, with the possible addition of suffixes, implemented in Prolog. The work described provides the framework and engine to interpret the D, N, and G stems along with accusative, dative and ventive endings.

## 1 Introduction

The goal of this research is to create a parser capable of taking a finite Akkadian verb, with some subset of direct-object, indirect-object, and ventive suffixes and returning the verb form, suffix form if applicable, and the radical stem. With the exception of the GUI, a Java wrapper running through a web-server, the program is written entirely in Prolog. The Definite Clause Grammar (DCG) rules that are used are an expansion of regular context-free grammars; they define a set of one or more expansions from a set of variables to a complete form. The program also functions as a generator of finite verbal forms, as the rules can both generate and recognize forms.

## 2 Parser design

The form to be parsed is input into the Prolog[1] interpreter as a character string of the finite verb form, comma delineated, and with long (macron) and elided (circumflex) vowels represented by double and triple vowels, respectively. This form, referred to as the Prolog Normal Form (PlNF), can either be input manually by the user or generated by the software from a character string input by the user. In the Prolog interpreter, backtracking is used to test for any possible finite form, with and without suffixes. When a matching form is found, a text string describing the form and the three radicals are returned, bound to variables to allow other programs to use this parsing engine. The parser is designed to handle any strong root and almost any weak root, but quadriliteral and doubly-weak roots are too irregular for a program of this scope. The program uses the ampersand (@)[2] to represent a generic weak radical, and simply the letters for W, Y, and N roots. In this manner weak roots are indicated in the form string as well as the three radicals, in most cases.

## 2.1 Vowels

Although vowels are quite important in Akkadian, in almost no cases are they needed to uniquely identify the stem. Due to this and the fact that knowing the vowel classes for each verb stem would limit the parser to those verbs which had been explicitly added to a dictionary, the program was designed from the beginning to ignore vowels when possible. Three main sets of rules were build, the `vs`, `vl`, and `vdl` for recognizing short, long, and doubly-long vowels respectively. These are used in rules when the verb, rather than the form, determines the vowel; in the others the vowels are hard-coded into the rules. As the program developed, needs were found for some other minor vowel rules, including the interchangeable 'a/e' for certain weak verbs and the interchangeable 'u/i' for some imperative forms. These allow only specified vowels, and not entire classes to be recognized. There are 22 vowel rules in the parser at the moment, but the chance that more are needed is quite small.

---

[2] The only other non-standard symbol needed was for the Shin, which is represented as the dollar sign ($).

The parser works on verbs that have been normalized, but the normalization process usually takes into account the form of the verb when determining vowel length from an inscription. This leads to a recursive problem that the parser works with normalized verbs to determine their forms; however the form must frequently be know in order to normalize the verb. This problem is partially ameliorated by the addition of wildcards to the vowel structure. Instead of forcing the user to know the vowel length, they can input the value and an asterisk and the program tests for all possible vowel lengths. As the value of the vowel would be known from an inscription, the decision was made not to allow a free wildcard, only to parse the three possible lengths of a given vowel. An example would be parsing the string `idda*k`, which gives responses from parsing *iddak* (G-preterite or durative), *iddäk* (N-preterite) and *iddâk* (N-durative).

## 2.2 Initial parsing functions

The central program is executed with a command, within the Prolog interpreter of the form:

```
?- verb(A, B, C, Type, [i,p,r,u,s]).
```
(Figure 1. Parsing a finite form, the A,B,C are bound to the radicals and Type to the returned string describing the type. The '?-' is the Prolog prompt.)

This calls the main 'verb' function, which then calls a parser for each stem twice, once with and once without the suffix interpreter. Each call begins by loading and compiling the appropriate source file, although in operation the compiled code is cached, thus not requiring multiple recompilation per session. This allows any new stem, as currently the parser only understands the G, D and N stems, to be added as a separate file with only two lines of code added to the main program. Each call to a stem parser then further breaks down the form, trying to parse it using one of the different tense parsers, i.e., D-durative or G-imperative, which then calls the various DCG rules to interpret strong and weak verb stems. All of the higher functions are standard Prolog declarations, all designed to call the proper DCG rules, which do the actual parsing/generation. These declarations, as they are called incrementally from `verb` can add, by concatenation, information to the type-string which is returned by the main program.

The `verb` function is then wrapped with a function, `akkadian` which takes as input the finite verb as a string and returns it parsed and normalized for each successful parse, taking care of the vowel wildcards and correctly separating the string into the PlNF for the `verb` function.

The nature of the DCG interpreter in Prolog allows it to parse a string to the end, or parse a prefix of that string and return the suffix. Each of the paradigm parsers is called twice, with the first pass forcing it to return no suffix and the second binding the suffix to a new variable. This bound variable is then passed to the suffix parser which can try combinations of Accusative, Dative and Ventive endings. If a suffix is successfully parsed, a string describing its type is returned and concatenated onto the verb type string. The parser is designed to ignore the conjunctive suffix –ma, but this currently is under development and does not always work.

As the program currently stands, the wrapper declarations are represented by 137 declarations and the suffix parser by 45 DCG rules.

## 2.3 Finite verb rules

The bulk of the work in the verb parsing, and programming, consists of the DCG rules which recognize the verb and bind the radicals to the variables (the A, B, and C in Fig. 1). It takes approximately 200 rules to parse a stem, which provides recognition of the preterite, durative, perfect, imperative, precative, and vetitive, in the first N, first W, active and stative, and first, second and third Aleph. As they are currently implemented, the forms are reduced to a base type, either explicitly, as in the `strnbasepret` rule for the N preterite, or implicitly by using other, previously declared forms. The modified forms are described using the rules for the base forms. An example is the plural G strong durative, where only one new rule was needed:

```
strgdur(Ca, Cb, Cc, [3,m,p]) -->
    strgdur(Ca, Cb, Cc, [3,c,s]), [uu].
strgdur(Ca, Cb, Cc, [3,f,p]) -->
    strgdur(Ca, Cb, Cc, [3,c,s]), [aa].
```

```
strgdur(Ca, Cb, Cc, [2,c,p]) -->
   strgdur(Ca, Cb, Cc, [2,m,s]), [aa].
strgdur(Ca, Cb, Cc, [1,c,p]) --> [n],
   [i], Ca, [a], Cb, Cb, vs, Cc.
```

(Figure 2. Plural G Durative rules. They represent, in normalized form: iparrusii[3], iparrusä, taparrusä and niparrus.)

In the DCG, a bracketed letter is a literal, a capitol letter is a variable and a lowercase another DCG, usually another form of the current verb or a vowel rule. The rule itself takes the form of: `name(argument[s])-->rule[s]`, where the arguments can be either defined in the rule(such as [2,c,p]) or variables to be bound by the rule itself; the rule is some string of bracketed literals, variables or another rule.

## 2.4 Generation

Within the Prolog interpretation of DCG rules, there is no defined sense of direction, so that the same rule-base, which can recognize a finite verb form, can also generate that form. Some of the more bizarre functions which handle string-to-PlNF conversion had to be rewritten, but fewer than five in the entire program. Due to the manner in which the vowels are handled (§ 2.1), multiple forms are usually generated, with different theme vowels. Although somewhat inconvenient, the correct form can easily be found with help of a dictionary. To generate a finite form using the low-level function, one calls the program with variables bound to the three radicals and the type, leaving the verb string unbound.

```
?- verb(p,r,s,['G', 'Preterite', 3, c,
s],Plnf).
```
(Figure 3. Generating a finite form, returned by the variable Plnf.)

In this case, the four values, *ipras*, *ipris*, *iprus*, *ipres*, are returned, but the correct *iprus* is easily identified by the user. Generation is most commonly not called at this low level, but through the `makeverb` wrapper function. This takes the three radicals for the verb and a list representing the type, and creates both a PlNF representation of the verb and a normalized string version. As this function was tuned to work with the Web interface, characters such as long vowels and the letter 'shin' are output as HTML codes.

The low-level generation, however, is quite useful for its ability to take wildcard commands, which currently cannot be input using the Web interface.

```
?- verb(p,r,s,['G', _, 3, c,
s],String).
```
(Figure 4. Generate any G-stem in the third, common singular form. The _ represents a non-binding variable.)

## 3  Interface

The user interface is a HTML form which then passes its input to a cgi-bin script which formats them properly for a Java wrapper to the Prolog interpreter.

### 3.1 HTML User Interface

The user interface consists of two HTML forms, one for parsing and one for generation, which are on the same page[4]. Both are quite simple, with the parser presenting the user with a text input area for the finite verb. The generator presents the user with a text area for inputting the stem, and selectors for the stem, tense and person.

The forms submit their data to a small UNIX shell script which formats them as arguments and passes them to the java parsing engine on the back-end processor. Due to the fact that parsing, especially with wildcard expansion, is computationally intensive, and Prolog is an interpreted language and therefore somewhat slow, the web-server, a PII-266 system, passes the actual work off to a backend, a dual Athlon 1900+ system. Computation time is indicated at the bottom of the page returned by the parser; it can process approximately ten forms per second.

### 3.2 Java Wrappers

The actual execution of the parsing from the web forms is done by a pair of java classes which interface between the shell script and the Prolog parser. Some code was needed to interpret the variables returned by the forms, and although Prolog code exists for this purpose, it was complex, and Java code to both interface with the script and the Prolog was readily available, making its selection easy. There are two Java classes, `parseAkk` and `akkadian`. The `akkadian` class does all of the work of interfacing with the Prolog, using the JPL[5] system and returning the HTML formatted answers or error messages, while the `parseAkk` class does the reformatting of the cgi-bin variables into useful Java types[6]. The akkadian class can be used as a command-line program for testing and quick parsing, but as it returns partial HTML, this use is no longer supported.

---

[3] Due to the difficulty of typesetting the macron usually used in representing the long vowel, the umlaut, (¨), will be used instead.

## 4 State of the program

The program is a work in progress, but this is the state as of April, 2002.

### 4.1 State and flow diagram

In the state diagram, the background color represents the code which is in the main program file (akk.pl) while the lighter blocks are the subordinate stem files (dstam.pl, nstem.pl and gstem.pl). Areas which are dashed are incomplete work in progress, and those with notes are incomplete pending more research to derive the forms.

The arrows in the flow diagram seem misleading, but every one of the verb rule-sets can have a suffix, so the joining of paths from the two different files is not incorrect.

The code can currently parse and generate based on all of the strong and singly-weak tri-radical G, D and N stems, with and without suffixes. Due to disagreements among grammarians, the second-weak D stem is parsed and generated correctly, but does not accept all of the grammatical variations given in the various grammars. When there is disagreement, the weight is given to Huehnergard's *Grammar*.

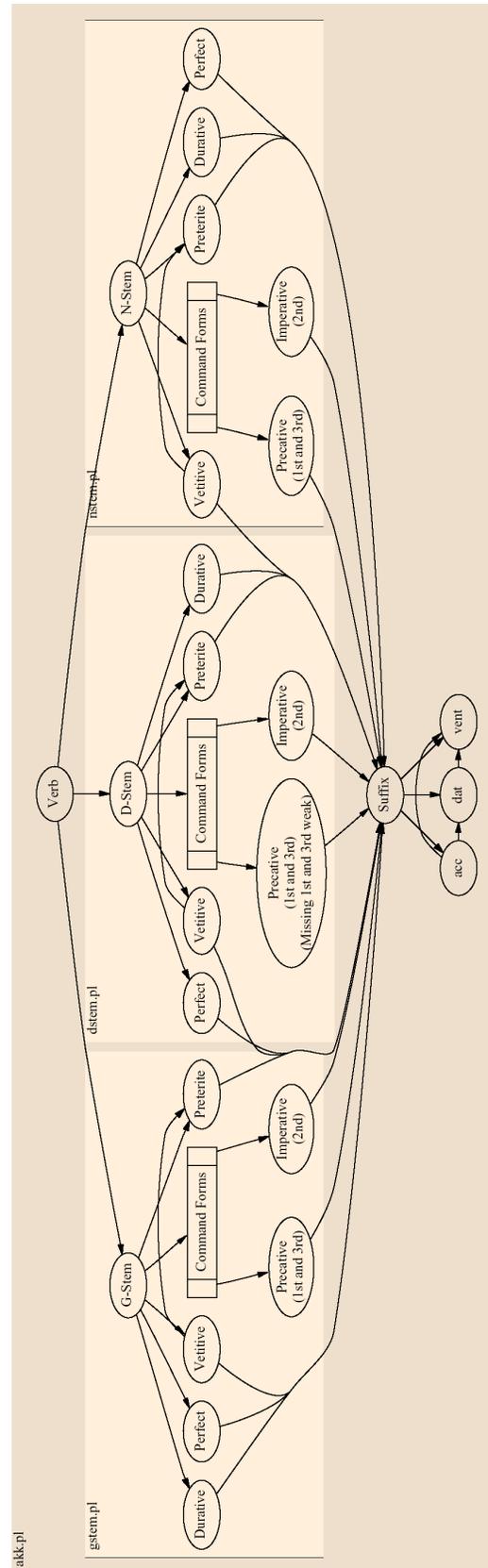

(Figure 5. The current flow of the program. Arrows represent the paths that the parser can take, in some cases indicating how rules are reused.)

---

## 4.2 Examples

Brief examples of the program parsing and generating finite verb forms. The is taken verbatim from the web interface. The examples for the G stem (Figures 6 and 7) are from the third weak infinitive *Qabû* (to speak), and are as conjugated in the Codex Hammurabi [Richardson, 2000]:

| Stem | q-b-@ |
|---|---|
| Parse | G Precative Third Weak 3 c s |
| Normalized form | liqbi |

(Figure 6. Parsing    liqbi: Third common singular G Precative[RICH, 126])

| Stem | q-b-@ |
|---|---|
| Parse | G Durative Third Weak 3 m p |
| Normalized form | iqabbuuuma |

(Figure 7. Parsing iqabbû-ma: Third masculine plural G Durative[RICH, 44], showing successful avoidance of the –ma suffix)

| Stem | n-d-@ |
|---|---|
| Parse | G Durative Third Weak 3 m p Accusative 3fs |
| Normalized form | inadduuu$i |

(Figure 8. Parsing innaddû$i: third weak G-durative, third masculine plural with a third feminine singular accusative ending [RICH, 74].)

| Stem | m-l-@ |
|---|---|
| Parse | D Preterite Third Weak 2 m s Ventive 2 f s |
| Normalized form | tumallinikkim |
| Stem | m-l-@ |
| Pare | D Preterite Third Weak 2 f s Ventive 2 f s |
| Normalized form | tumalliiinikkim |
| Stem | m-l-@ |
| Parse | D Durative Third Weak 2 f s Ventive 2 f s |
| Normalized form | tumalliiinikkim |

(Figure 9.    Parsing tumalli*nikkim, showing vowel wildcards. It parses into a third weak D-preterite, second masculine singular with a short 'i' Third Weak D-durative or D-preterite second feminine singular, with a double-long 'i', all with second feminine singular ventive ending. The system took 0.473 seconds to parse all the forms.)

An important note to figure 9, a form in Akkadian can arise from two convergent streams of conjugation, and in this case the verb form *tumallînimakkim* can be either preterite or durative.

## 5    Conclusion and future work

The idea behind this research was not to simulate the evolution of Akkadian, either historically as is done in comparative Semitic studies, or to use some sort of theoretical, two-stage morphological rules as in [Kataja, 1988]. Rather the goal was to create a system which could be used by researchers and students to help in understanding Akkadian texts. The complexities and many weak variants of the verb stems prevent the two-stage rules from ever yielding a useful parser, and there are many unneeded complexities in the historical evolution. This system as it evolved is, perhaps, needlessly intricate, but it works quickly and well and adding new verb stems can be done in about 8 hours of work.

Work continues on the parser, in teaching Assyriologists to use it in their daily work, in checking for errors in the rules, and in adding more verbal forms. Currently the effort is directed to add the $-stem, the last of the simple stems, to the parser and teach members of the Department of Near Eastern and Judaic Studies at Brandeis University to allow them to make use of the program. Future plans begin with work to add wildcards to the generation part of the web interface, to allow the generation of, for instance, all G preterite forms. This would, among other thing, aid in testing the accuracy of the generated forms, and by extension, the parser itself. They also include adding some of the more common derived stems, from each of the four simple stems there are three derived stems, a 't', an 'n' and a 'tn', for example, the G yields the Gt, Gn and Gtn. Suffix generation needs to be added to the Web interface and along with that some assimilation rules should be added. In the language, if a finite form ending with a semi-weak is followed by a suffix beginning with a Shin, the weak assimilates, for example 'm$' becomes '$$'.

## 6    Acknowledgements

The author would like to thank Professor James Pustejovsky for his assistance and guidance, Professor Jacques Cohen for his advice on Prolog and Professors David Wright, Tzvi Abusch and Dr. Kathryn Kravitz of the Department of Near Eastern and Judaic Studies at Brandeis University for their assistance with Akkadian grammar.